\documentclass[10pt,twocolumn,letterpaper]{article}

\usepackage{iccv}
\usepackage{times}
\usepackage{epsfig}
\usepackage{graphicx}
\usepackage{amsmath}
\usepackage{amssymb}

\usepackage{times}
\usepackage{epsfig}
\usepackage{graphicx}
\usepackage{amsmath}
\usepackage{amssymb}

\usepackage{float}
\usepackage[caption=false]{subfig}

\newcommand{\tablestyle}[2]{\setlength{\tabcolsep}{#1}\renewcommand{\arraystretch}{#2}\centering\footnotesize}

\usepackage{pifont}         % http://ctan.org/pkg/pifont
\newcommand{\cmark}{\ding{51}}%
\newcommand{\xmark}{\ding{55}}%

\newcommand{\red}{\textcolor[rgb]{0,0,0}}

\usepackage{multicol,multirow}
\newlength\savewidth\newcommand\shline{\noalign{\global\savewidth\arrayrulewidth
		\global\arrayrulewidth 1pt}\hline\noalign{\global\arrayrulewidth\savewidth}}

\def\eg{\emph{e.g}\onedot} 
\def\ie{\emph{i.e}\onedot} 

\usepackage[breaklinks=true,bookmarks=false]{hyperref}

\iccvfinalcopy % *** Uncomment this line for the final submission

\def\iccvPaperID{****} % *** Enter the ICCV Paper ID here

\ificcvfinal\fi

\begin{document}

%%%%%%%%% TITLE
\title{Cross-Modal Attention Consistency for Video-Audio Unsupervised Learning}

\author{Shaobo Min\textsuperscript{1}\thanks{Work was done during internship at Microsoft.}, Qi Dai\textsuperscript{2}, Hongtao Xie\textsuperscript{1}\thanks{Corresponding author.}, Chuang Gan\textsuperscript{3}, Yongdong Zhang\textsuperscript{1}, and Jingdong Wang\textsuperscript{2}\\
\textsuperscript{1}University of Science and Technology of China\\
\textsuperscript{2}Microsoft Research Asia\\
\textsuperscript{3}MIT-IBM Watson AI Lab\\
%{\tt\small \{mbobo,cq14\}@mail.ustc.edu.cn, hantao.yao@nlpr.ia.ac.cn, \{htxie,zhazj,zhyd73\}@ustc.edu.cn}
}
\maketitle
\ificcvfinal\thispagestyle{empty}\fi

%%%%%%%%% ABSTRACT
\begin{abstract}
 Cross-modal correlation provides an inherent supervision for video unsupervised representation learning.
%  and has shown to be more effective than single-modality learning. 
 Existing methods focus on distinguishing different video clips by visual and audio representations.
%  while ignoring the local correspondence between spatio-temporal regions and acoustic frequencies. 
 We human visual perception could attend to regions where sounds are made, and our auditory perception could also ground their frequencies of sounding objects,
 which we call bidirectional local correspondence.  
 Such supervision is intuitive but not well explored
 in the contrastive learning framework.
  This paper introduces a  pretext task,
  Cross-Modal Attention Consistency (CMAC),
  for exploring the bidirectional local correspondence property.
  The CMAC approach aims to align the regional attention generated purely from the visual signal
  with the target attention generated under the guidance of acoustic signal,
  and do a similar alignment for frequency grounding on the acoustic attention.
  Accompanied by a remoulded cross-modal contrastive loss
  where we consider additional within-modal interactions, the CMAC approach works effectively for enforcing 
  the bidirectional alignment.
  Extensive experiments on six downstream benchmarks demonstrate that CMAC can improve the state-of-the-art performance on both visual and audio modalities.
\end{abstract}

%%%%%%%%% BODY TEXT\section{Introduction}
\section{Introduction}
Unsupervised image representation learning \cite{wang2015unsupervised,pathak2017learning,gidaris2018unsupervised,he2020momentum,chen2020simple} has attracted great attention recently, which attempts to learn useful knowledge from massive unlabeled data and transfer to various downstream tasks.
%Without human annotations, these methods design various proxy tasks, such as predicting the rotation angle \cite{} and Jigsaw order \cite{}, to supervise the visual encoder.
Nevertheless, the efforts in video counterpart are still inapparent, though the multi-modal nature is particularly suitable for unsupervised learning by providing intensive coherence and variation information.
%In reality, multi-modal information dominate the human sensor \cite{}, such as visual and sound signals.
This paper focuses on the Video-Audio Unsupervised Learning task \cite{korbar2018cooperative,morgado2020audio,patrick2020multi,korbar2018cooperative}, which aims at exploring cross-modal clues to simultaneously learn visual and audio representations from massive unlabeled videos. 

\begin{figure}[h]
    \centering
    \includegraphics[width=1\columnwidth]{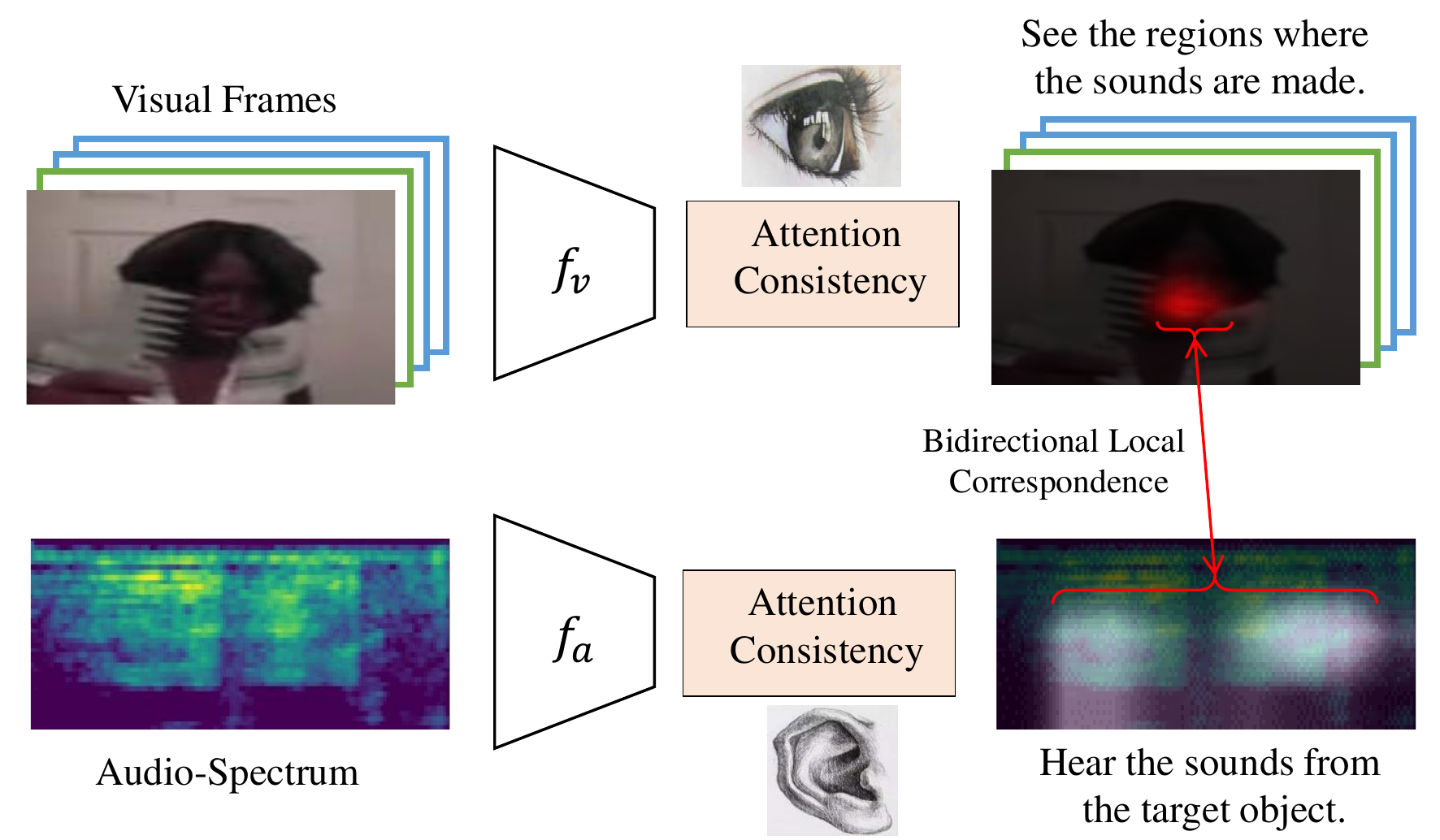}
    \caption{CMAC expects the visual encoder to focus on the regions where sounds are made and the audio encoder to focus on sounds from the interested objects. }
    \label{fig:intro}
    \vspace{-0.19cm}
\end{figure}

A general technique \cite{korbar2018cooperative,morgado2020audio} in the literature is to leverage the intrinsic correspondence between visual frames and audio waves by determining whether they are from the same video instance.
With recent advances of contrastive learning \cite{wu2018unsupervised,he2020momentum}, these approaches make positive (negative) video-audio pairs similar (dissimilar).
%Specifically, the contrastive learning \cite{wu2018unsupervised,he2020momentum} is widely used to attract positive visual-audio representations from the same video and repel negative visual-audio pairs from different videos.
%global representations from visual and audio encoders for different visual-audio pairs.
Based on this paradigm, AVTS \cite{korbar2018cooperative} designs a harder synchronization supervision, which defines positive visual-audio pairs only when they are from the identical clip of a video.
%visual and audio representations as positive pair only when they are from the same clip of a video instance.
%Compared with distinguishing video instances, AVTS further requires visual and audio encoders to determine whether an input visual-audio pair is synchronous.
%AVSlowFast \cite{xiao2020audiovisual} proposes a visual-audio encode architecture that contains slow and fast visual pathways to explore knowledge at different temporal scales.
In addition to cross-modal discrimination, AVID \cite{morgado2020audio} considers within-modal discrimination, which simultaneously explores the relationship between visual-audio, visual-visual, and audio-audio pairs.
Recently, GDT \cite{patrick2020multi} develops a unified visual-audio contrastive framework by exploring more comprehensive data augmentations and modality relationship.

Despite the impressive performance of the above methods, their pretext supervisions only consider the instance level relationship between global modality representations, \eg the aggregated spatio-temporal features and audio spectrogram features \cite{alwassel2019self,patrick2020multi}.
%which are obtained by summing up local features along temporal-spatial and audio-spectrum dimensions \cite{}.
This inevitably discards fine-grained local clues, \ie the relation between each spatio-temporal region and foreground sound frequency.
In the human sensory system, our visual perception is sensitive to the spatial regions where sounds are made, and vice versa our auditory perception is sensitive to the sound frequencies belonging to foreground objects.
This phenomenon, called \emph{bidirectional local correspondence}, is natural and intuitive to associate spatio-temporal visual clues with acoustic frequencies.
%Such dense local correspondence is natural and intuitive that can associate spatio-temporal visual clues with acoustic frequencies.
\red{Some early works \cite{arandjelovic2017look,owens2016ambient,owens2018audio} study the global alignment between modalities for representation learning, while ignoring such local relation.
%signals for representation learning by matching the sound with corresponding image.
One recent exploration \cite{hu2019deep} considers it in image feature learning via co-clustering. However, their method has not been validated in the challenging large-scale video unsupervised learning task.}
%Unfortunately, existing video-audio unsupervised methods lack a specific mechanism to capture such information between visual and acoustic signals in an unsupervised manner.
%such a dense correspondence between temporal-spatial visual clues and audio-spectrum signals in an unsupervised manner.

In this paper, we propose a novel pretext task, namely Cross-Modal Attention Consistency (CMAC), for exploring the \emph{bidirectional local correspondence} property between visual and acoustic signals.
%Cross-Modal Attention Consistency (CMAC) method that introduces a new pretext task by exploring the local correspondence patterns between visual and acoustic signals.
%that captures the local correspondence patterns between visual and acoustic signals as a new pretext supervision for visual-audio unsupervised learning.
%intrinsic correspondence between spatio-temporal visual clues and audio-spectrum signals as visual-audio supervision.
The core insight is to make the visual encoder attend to regions where sounds are made and the audio encoder attend to sound frequencies from interested objects, as shown in Fig. \ref{fig:intro}.
To achieve the goal, CMAC proposes to align the \emph{regional attention} generated purely from the visual signal with the \emph{target attention} generated under the guidance of acoustic signal, and do a similar alignment for frequency grounding on the acoustic attention.
Compared with traditional cross-modal contrastive manner \cite{alwassel2019self,patrick2020multi}, CMAC provides a novel mechanism that considers bidirectional local correspondence between spatio-temporal visual clues and audio-spectrogram signals via attention consistency.

%To address the above issue, CMAC aligns the \emph{regional attention} generated purely from the visual signal with the \emph{target attention} generated under the guidance of acoustic signal, and do a similar alignment for frequency grounding on the acoustic attention.
%CMAC devises a pyramid attention mechanism to model the local correspondence, which highlights the interested parts in one modality with guidance from the other modality.

%With this inspiration, the main challenge of CMAC is how to localize visual regions and sound frequencies without human annotations.

Specifically, to produce the target cross-modal attention without human annotations, we devise a pyramid attention mechanism.
CMAC first dynamically learns a set of adaptive filters (kernels) in each modality.
To generate target attention for one modality (\eg visual), the learned filters from the other modality (\ie audio-induced) are utilized to perform filtering on its representations (\ie visual).
The predicted audio-guided visual attention map thus indicates the spatio-temporal regions that are most related to audio signals, and vice versa for the visual-guided audio attention map. 
With the learned cross-modal attention maps as supervision, CMAC then preserves the attention consistency between them and the single-modality induced attention maps. 
%Compared with traditional cross-modal contrastive manner \cite{alwassel2019self,patrick2020multi}, CMAC provides a novel mechanism that considers bidirectional local correspondence between spatio-temporal visual clues and audio-spectrum signals via attention consistency.

The proposed attention consistency works effectively in accompany with a remoulded contrastive loss, where additional within-modal information is considered.
%enriches the representations with both within- and cross-modal information.
%As a complementary, CMAC also leverages a remoulded contrastive loss to enrich features with both within- and cross-modal information.
Different from AVID \cite{morgado2020audio}, we only involve within-modal negative samples from previous batch data, thereby without extra memory cost for resampling within-modal positive samples for the current batch.
This simple yet effective modification successfully boosts the performance.
%The loss brings two merits: a) introducing within- and cross-modal instance discrimination knowledge, and b) bridging the modality gap in the adaptive filters.
Experiments on various downstream tasks including action recognition, video retrieval, and audio classification show that CMAC can improve the state-of-the-art results.

%To address the above issue, CMAC leverages the correlation filter between modalities to localize local maximum response.
%Specifically, CMAC first generates a set of filters from respective audio and visual representations.
%Due to modality gap, the audio-induced filters cannot be directly applied to temporal-spatial visual representations and vice versa.
%To this end, a modality transformation module is then designed to project theses filters into a joint embedding space, where visual-induced and audio-induced filters from the same video instance are aligned via contrastive learning.
%This enables audio-induced filters to capture some visual semantics that are related to audio signals and vice versa.
%Finally, the audio- and visual-induced filters are applied to respective temporal-spatial visual representations and audio-spectrum representations via correlation filtering.
%The generated response maps of two modalities indicate temporal-spatial regions and sound-frequency that are most related to audio-frequency and visual objects, respectively.

The contributions of this paper are three-fold: (1) We show that the \emph{bidirectional local correspondence} is effective in cross-modal representation learning; (2) A novel pretext task namely Cross-Modal Attention Consistency (CMAC) is introduced for unsupervised learning; (3) New state-of-the-art results are achieved by CMAC on various downstream benchmarks.

\section{Related Works}
Unsupervised representation learning targets at learning good data representations in an unsupervised manner.
%Two kinds of unsupervised learning methods are most related to this paper which are: unsupervised learning from images and cross-modal unsupervised learning.

\noindent\textbf{Unsupervised learning from images.} Due to unavailable human annotation, existing methods focus on designing pretext tasks \cite{doersch2015unsupervised,wang2015unsupervised,pathak2017learning,gidaris2018unsupervised} to learn good data representations.
For example, a seminal work \cite{dosovitskiy2014discriminative} augments each image into several transformations and regards each image as an exemplar, thus a traditional classification task can be defined as recognizing image transformations into their exemplar category.
Besides, CFN~\cite{noroozi2016unsupervised} designs a jigsaw puzzle task by predicting orders of different patches from the same image, and RotNet~\cite{gidaris2018unsupervised} predicts the rotation angles of an image.
These proxy tasks aim to understand some low-level image concepts, such as rotation invariance and spatial relationship among image patches.
Recently, contrastive loss \cite{hadsell2006dimensionality,wu2018unsupervised,oord2018representation,hjelm2018learning,zhuang2019local,henaff2019data,caron2020unsupervised,grill2020bootstrap} has attracted great attention, which makes different transformations from the same image attract and those from different images repel.
Compared to fixed prototypes in exemplar methods, the contrastive loss can vary on-the-fly training.
Two representative methods are MoCo \cite{he2020momentum} and SimCLR \cite{chen2020simple}.
MoCo designs a dynamic dictionary with a moving-averaged encoder to store a large-scale and consistent instance representation set, and SimCLR explores the importance of projection head and comprehensive data augmentations in unsupervised learning.

\begin{figure*}[h]
    \centering
    \includegraphics[width=1.9\columnwidth]{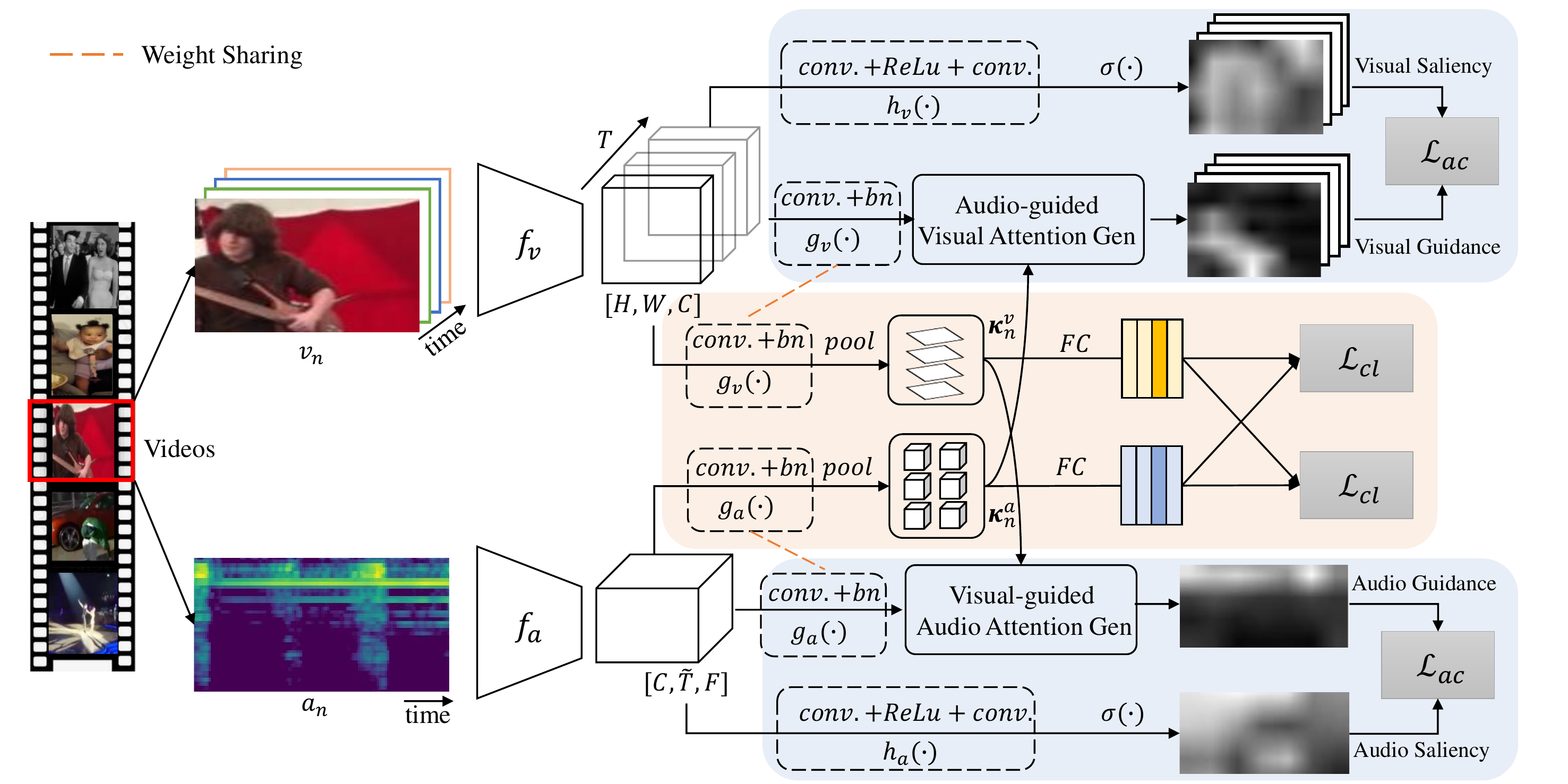}
    \vspace{0.1cm}
    \caption{\small Overview of CMAC. 
    We aim to align \emph{regional attention} generated purely from the visual signal with \emph{target attention} generated under the guidance of audio signal, and do the similar alignment for the audio counterpart.
    CMAC first produces filters for both modalities, and then generates cross-modal guided attention by searching the most matched patterns in one modality in terms of filters from the other modality.
    The guided attention is used to supervise the single-modal attention via attention consistency $\mathcal{L}_{ac}$.
    A remoulded contrastive loss $\mathcal{L}_{cl}$ is leveraged to bridge the semantic gap between filters of two modalities by considering additional within-modal negative samples. 
    }
    \label{fig:cmac}
    %\vspace{-0.4cm}
\end{figure*}

\noindent\textbf{Cross-modal unsupervised learning.} Compared to images, videos are more commonly seen and contain a rich variety of modalities,~\emph{e.g.} visual, audio, and speech signals.
The correlation between different modalities is a natural supervision that has attracted increasing interest \cite{de1994learning}.
Some works \cite{sun2019videobert,miech2020end,li2020learning,nagrani2020speech2action} leverage the speech as a weak supervision, and other works explore the audio self-supervision to boost visual localization \cite{senocak2018learning}, audio-visual source separation \cite{harwath2018jointly,gao2018learning,gao2019co,rouditchenko2019self}, and representation learning \cite{arandjelovic2017look,owens2018audio,piergiovanni2020evolving,arandjelovic2018objects}.
In this work, we focus on visual-audio unsupervised learning, which aims at jointly learning visual and audio representations in an unsupervised manner.
A general paradigm is to detect whether the visual and audio signals are from the same video, which is commonly solved via cross-modal contrastive learning \cite{korbar2018cooperative,morgado2020audio,patrick2020multi,xiao2020audiovisual}.
AVTS \cite{korbar2018cooperative} further employs the temporal synchronization by defining positive visual-audio pairs only when they are temporally synchronized, constraining the representations to understand the temporal content.
In addition to the cross-modal learning, AVID \cite{morgado2020audio} introduces the within-modal information via extra visual-visual and audio-audio contrastive learning, but it requires doubling the memory cost for both within- and cross-modal positive pairs.
GDT \cite{patrick2020multi} explores more comprehensive data augmentations and data pair constructions.
Different from cross-modal contrastive learning, XDC \cite{alwassel2019self} leverages unsupervised clustering in one modality to supervise representations of the other modality. AVSlowFast \cite{xiao2020audiovisual} proposes a visual-audio encode architecture that contains slow and fast visual pathways to explore knowledge at different temporal scales.
Though above methods achieve promising results, they only consider cross-modal instance discrimination by distinguishing different clips, while ignoring the local correspondence between spatio-temporal regions and acoustic frequencies.

\red{Notably, the visual-audio local correspondence has been partially studied in visual-audio localization \cite{zhao2018sound,rouditchenko2019self} and representation learning \cite{hu2019deep} methods.
For example, DMC \cite{hu2019deep} treats the local features in image feature maps as a set of distinct components, and learns the visual and auditory subnets by co-clustering them.
Though the learned visual features perform well in image classification on small datasets, however, this method has not been validated in the challenging unsupervised video pretraining task.
In contrast to the clustering-based learning in DMC, in this paper we propose a novel attention-based method to explore the effectiveness of local correspondence in this challenging task, rather than learning exact localization between modalities.}

%For example, DMC \cite{hu2019deep} treats the local features in image feature maps as a set of distinct components, and it aims to learn the correspondence between local features of different modalities by co-clustering them with image-audio pairs.
%Different from these methods that focus on how to better localize sound regions in an unsupervised manner, this paper explores how to utilize such a local correspondence as a supervision to guide video-audio representation learning.
%Interestingly, these methods could be a component of CMAC to learn a better correspondence supervision.

\section{Method}
\subsection{Preliminaries}
Suppose we have a set of videos $\{\boldsymbol{x}\}$ consisting of a visual track (RGB frames) and an audio track (sound), the target of visual-audio unsupervised learning is to learn feature encodings $f_v(\cdot)$ and $f_a(\cdot)$ for both modalities that can well transfer to various downstream visual or audio tasks.
%respective visual and audio representation encoding that can well transfer to various downstream video or audio tasks.

Formally, we define $(\boldsymbol{v}_{n}, \boldsymbol{a}_{n})$ as the encoded visual and audio representations of the $n$-th video, \ie $\boldsymbol{v}_{n}=f_v(\boldsymbol{x}_n)$, $\boldsymbol{a}_{n}=f_a(\boldsymbol{x}_n)$.
A general paradigm is to detect synchronized pair of $(\boldsymbol{v}$, $\boldsymbol{a})$ in a contrastive manner.
When $\boldsymbol{v}$ and $\boldsymbol{a}$ are from the same location of one video, they are synchronized and identified as positive sample pair.
When they are from different videos, they form a negative pair.
%if $(\boldsymbol{v}$ and $\boldsymbol{a})$ are from different videos
By sampling massive positive and negative visual-audio pairs, the \emph{noise-contrastive loss} \cite{chen2020simple,wu2018unsupervised} can be formulated as:
\begin{eqnarray}\label{eq:L_nce}
\mathcal{L}_{nce}(\boldsymbol{v},\boldsymbol{a}) = - \frac{1}{N}\sum_{n=1}^{N}log \frac{sim(\boldsymbol{v}_{n},\boldsymbol{a}_{n})}{\sum_{m}sim(\boldsymbol{v}_{n},\boldsymbol{a}_{m})},
\end{eqnarray}
where $sim(x,y)=e^{<x,y>_{\tau}}$, and $<\cdot,\cdot>_{\tau}$ is the cosine similarity divided by a temperature parameter $\tau$. 
$N$ is the size of batch.
When $m\neq n$, $\boldsymbol{a}_{m}$ is a negative sample.
For simplicity, we omitted the visual and audio augmentations here.
The final cross-modal contrastive loss becomes:
\begin{eqnarray}\label{eq:L_cmcl}
 \mathcal{L}_{nce}(\boldsymbol{v},\boldsymbol{a}) +\mathcal{L}_{nce}(\boldsymbol{a},\boldsymbol{v}).
\end{eqnarray}
Eq.~\eqref{eq:L_cmcl} indeed leverages the intrinsic synchronization between visual and audio signals as supervision,~\emph{i.e.,} detecting whether the input visual and audio representations are synchronized.
However, in human perception mechanism, a more intuitive supervision is that our visual system usually focuses on the regions where sounds are made, and our auditory system focuses on the frequencies that belong to interested objects, called \emph{bidirectional local correspondence}.
To introduce such an important pretext supervision, we propose a novel Cross-Modal Attention Consistency framework, of which the illustration is shown in Fig.~\ref{fig:cmac}.

\subsection{Cross-Modal Attention Consistency}
The proposed Cross-Modal Attention Consistency (CMAC) aims to explore the bidirectional local correspondences between $\boldsymbol{v}$ and $\boldsymbol{a}$ to supervise the visual encoder $f_v(\cdot)$ and audio encoder $f_a(\cdot)$.
%constrain $f_v(\cdot)$ to focus on the region where the sounds are made and $f_a(\cdot)$ to focus on the acoustic frequencies belonging to the salient object.
To achieve the goal, CMAC tackles two challenges: a) how to localize the crucial spatio-temporal regions and acoustic frequencies as supervisions, and b) how to constrain $f_v(\cdot)$ and $f_a(\cdot)$ to concentrate on the localized regions and frequencies.

Given a synchronized pair\footnote{$\{C,T,H,W\}$ denote channel, frame, height, and width respectively. $\{\widetilde{T},F\}$ indicate the time and frequency. $\widetilde{T}$ is determined by window width of FFT and is generally different from $T$.} $\boldsymbol{v}_{n}\in \mathbb{R}^{C\times T\times H\times W}$ and $\boldsymbol{a}_{n}\in \mathbb{R}^{C\times \widetilde{T}\times F}$, CMAC first estimates a set of adaptive filters (kernels) on visual and audio representations by
\begin{eqnarray}\label{eq:kenel}
\boldsymbol{\kappa}_{n}^{v} = pool(g_{v}(\boldsymbol{v}_{n})), \quad \boldsymbol{\kappa}_{n}^{a} = pool(g_{a}(\boldsymbol{a}_{n})),
\end{eqnarray}
where $g_v(\cdot)$ and $g_a(\cdot)$ are two transformation functions implemented by \emph{conv.}~+~\emph{bn} operations.
%convolutional functions that transform $\boldsymbol{v}_{n}$ and $\boldsymbol{a}_{n}$ into a set of filters.
$\boldsymbol{\kappa}_{n}^{v}$ and $\boldsymbol{\kappa}_{n}^{a}$ are the produced filters for visual and audio modalities, respectively.
$pool(\cdot)$ is the pooling function that controls the kernel size of $\boldsymbol{\kappa}_{n}^{v}$ and $\boldsymbol{\kappa}_{n}^{a}$.
Here, we adopt the global average pooling for simplicity, which means $\boldsymbol{\kappa}_{n}^{v}\in \mathbb{R}^{C\times 1\times 1\times 1}$ and $\boldsymbol{\kappa}_{n}^{a}\in \mathbb{R}^{C\times 1\times 1}$.
These filters provide the modality-specific essentials for capturing related contents in the other modality.
With $\boldsymbol{\kappa}_{n}^{v}$ and $\boldsymbol{\kappa}_{n}^{a}$, we leverage the attention mechanism to highlight the corresponding contents by devising a \emph{pyramid correlation filtering} module for each modality.

%To localize temporal-spatial regions related to the target sounds, a straightforward strategy is to filter $\boldsymbol{v}_{n}$ with the audio kernel $\boldsymbol{\kappa}_{n}^{a}$ along temporal-spatial dimensions.
%However, $\boldsymbol{v}_{n}$ and $\boldsymbol{\kappa}_{n}^{a}$ are from different modalities with serious statistical gap, which cannot be directly interacted.

\noindent\textbf{Pyramid Correlation Filtering.} 
To generate the target attention map for one modality, \eg visual, we utilize the learned filters from the other modality, \ie audio, to perform Pyramid Correlation Filtering (PCF) on its representations (\ie visual).
%With cross-modal filters $\boldsymbol{\kappa}_{n}^{v}$ and $\boldsymbol{\kappa}_{n}^{a}$, a pyramid correlation filtering (PCF) is designed to localize temporal-spatial regions in $\boldsymbol{v}_{n}$ and acoustic frequencies in $\boldsymbol{a}_{n}$.
The core insight is to search the most matched patterns in one modality in terms of filters from the other modality.
Consequently, the attention map $\boldsymbol{s}_{n}^{v}$ ($\boldsymbol{s}_{n}^{a}$) is calculated by convolving the filter $\boldsymbol{\kappa}_{n}^{a}$ ($\boldsymbol{\kappa}_{n}^{v}$) over the modality representation $\boldsymbol{v}_{n}$ ($\boldsymbol{a}_{n}$), which can be formulated as
%correlation filtering is used to calculate a response map from modality representations and cross-modal filters by:
\begin{eqnarray}\label{eq:pcf}
\boldsymbol{s}_{n}^{v} = norm(\boldsymbol{\kappa}_{n}^{a}*g_{v}(\boldsymbol{v}_{n})),~ \boldsymbol{s}_{n}^{a} = norm(\boldsymbol{\kappa}_{n}^{v}*g_{a}(\boldsymbol{a}_{n})),
\end{eqnarray}
where $\boldsymbol{s}_{n}^{v}\in \mathbb{R}^{T\times H\times W}$ is the audio-guided visual attention for $\boldsymbol{v}_{n}$, and $\boldsymbol{s}_{n}^{a}\in \mathbb{R}^{\widetilde{T}\times F}$ is the visual-guided audio attention for $\boldsymbol{a}_{n}$.
Here $g_{v}(\cdot)$ and $g_a(\cdot)$ are shared with that in Eq. (\ref{eq:kenel}).
%In Eq.~\eqref{eq:pcf}, $\boldsymbol{v}_{n}$ and $\boldsymbol{a}_{n}$ have been transformed via two projection heads, while we omit the functions for simplicity.
$(*)$ denotes the convolution operation.
For instance, $\boldsymbol{s}_{n}^{v}$ measures the similarity response between $\boldsymbol{\kappa}_{n}^{a}$ and each local feature in $\boldsymbol{v}_{n}$.
%Here, we adopt cosine distance to measure the similarity.
$norm(\cdot)$ maps the response map into $[0,1]$, where the cosine-based normaliztion is used in this paper.
%which uses linear scale in this paper.
The predicted audio-guided visual attention $\boldsymbol{s}_{n}^{v}$ indicates the spatio-temporal regions that are most related to the audio signals, \ie regions where the sounds are made.
%which temporal-spatial regions are most related to the target audio signal,~\emph{i.e.,} regions where the sounds are made.
\begin{figure}[h]
    \centering
    \includegraphics[width=1\columnwidth]{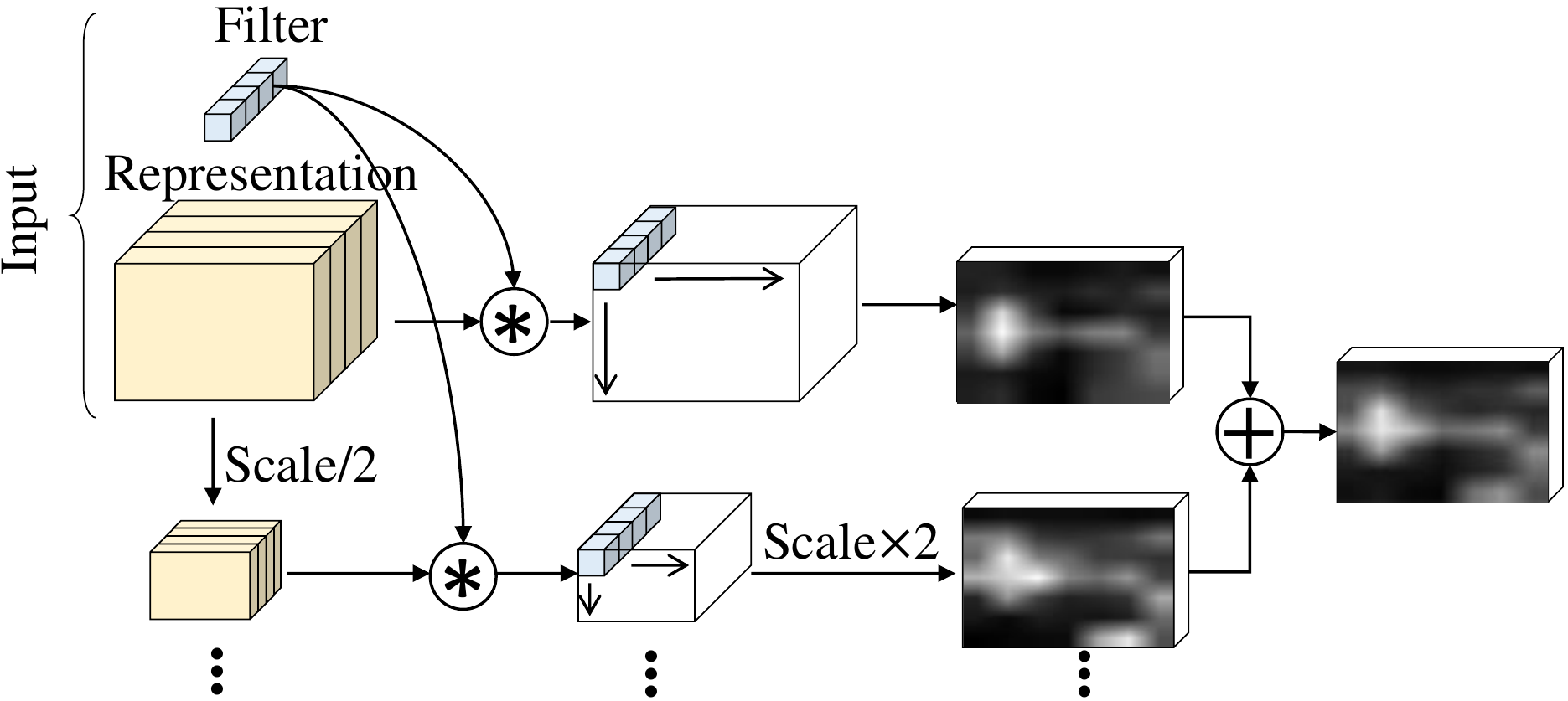}
    \caption{Pyramid Correlation Filtering (PCF). The inputs of PCF contain a representation and corresponding filters. PCF convolves filters over the representations at each position and obtains a response map. Notably, a pyramid scaling strategy is used to fuse the multi-scale responses. }
    \label{fig:pcf}
\end{figure}
Similarly, the visual-guided audio attention $\boldsymbol{s}_{n}^{a}$ indicates the acoustic frequencies from the interested objects.
%which acoustic frequencies are most related to the target visual signals,~\emph{i.e.,} frequencies from the interested objects.

It is worth noting that we adopt a pyramid scaling strategy to obtain better attention clues.
Since both $\boldsymbol{s}_{n}^{v}$ and $\boldsymbol{s}_{n}^{a}$ preserve the original scales of representations, we further downsample $\boldsymbol{v}_{n}$ and $\boldsymbol{a}_{n}$ to the half resolution and calculate the filter response again, as shown in Fig.~\ref{fig:pcf}.
%Notably, both $\boldsymbol{s}_{n}^{v}$ and $\boldsymbol{s}_{n}^{a}$ are obtained at original representation scale.
%Thus, we adopt a pyramid scale strategy by downsampling $\boldsymbol{v}_{n}$ and $\boldsymbol{a}_{n}$ to half resolution, and calculate the filter response again at a half scale, as shown in Figure~\ref{fig:cmac}.
As a result, we fuse the response maps at different scales to generate the final attention map.

\noindent\textbf{Attention Consistency.} For exploring local corresponding patterns between $\boldsymbol{v}_{n}$ and $\boldsymbol{a}_{n}$, the learned $\boldsymbol{s}_{n}^{v}$ and $\boldsymbol{s}_{n}^{a}$ can be regarded as pseudo labels that $f_v(\cdot)$ and $f_a(\cdot)$ should focus on.
As illustrated in Fig.~\ref{fig:cmac}, CMAC first exploits two saliency detection heads to directly infer the within-modal attention maps $\hat{\boldsymbol{s}}_{n}^{v}$, $\hat{\boldsymbol{s}}_{n}^{a}$ purely from $\boldsymbol{v}_{n}$, $\boldsymbol{a}_{n}$ by:
\begin{eqnarray}\label{eq:s}
\hat{\boldsymbol{s}}_{n}^{v} = \sigma(h_v(\boldsymbol{v}_{n})), \quad \hat{\boldsymbol{s}}_{n}^{a} = \sigma(h_a(\boldsymbol{a}_{n})),
\end{eqnarray}
where $h_v(\cdot)$ and $h_a(\cdot)$ are two convolution blocks for predicting the attention maps, and $\sigma(\cdot)$ is the sigmoid function.
The learned single-modality induced maps $\hat{\boldsymbol{s}}_{n}^{v}$, $\hat{\boldsymbol{s}}_{n}^{a}$ imply the concentrations of the modality encoders $f_v(\cdot)$, $f_a(\cdot)$.
Accordingly, the attention consistency can be preserved by aligning them with the previous cross-modal attention maps $\boldsymbol{s}_{n}^{v}$, $\boldsymbol{s}_{n}^{a}$ as follows:
%we propose to preserve the attention consistency between them and the previous cross-modal attention maps $\boldsymbol{s}_{n}^{v}$, $\boldsymbol{s}_{n}^{a}$ as follows:
%Since $\hat{\boldsymbol{s}}_{n}^{v}$ and $\hat{\boldsymbol{s}}_{n}^{a}$ are directly inferred from within-modal representations, they indicate the focuses of $f_v(\cdot)$ and $f_a(\cdot)$.
%Thus, $\hat{\boldsymbol{s}}_{i}^{v}$ and $\boldsymbol{s}_{i}^{v}$ are constrained to be consistent by:
\begin{equation}\label{eq:sc}\small
\mathcal{L}_{ac}(\boldsymbol{s}_{n}^{v},\hat{\boldsymbol{s}}_{n}^{v})=||{\boldsymbol{s}}_{n}^{v}-\hat{\boldsymbol{s}}_{n}^{v}||_{2}^{2},~~
\mathcal{L}_{ac}(\boldsymbol{s}_{n}^{a},\hat{\boldsymbol{s}}_{n}^{a})=||{\boldsymbol{s}}_{n}^{a}-\hat{\boldsymbol{s}}_{n}^{a}||_{2}^{2}.
\end{equation}
%and similar $\mathcal{L}_{ac}(\boldsymbol{s}_{i}^{a},\hat{\boldsymbol{s}}_{i}^{a})$ is used.

\noindent\textbf{Contrastive Learning.} 
In the above Pyramid Correlation Filtering, directly performing the cross-modal correlation filtering would lead to degradation as the filter and representation are from different modalities with a serious statistical gap.
To this end, we leverage the contrastive learning to bridge the modality gap between $\boldsymbol{\kappa}_{n}^{a}$, $\boldsymbol{\kappa}_{n}^{v}$ and $\boldsymbol{v}_{n}$, $\boldsymbol{a}_{n}$.
Specifically, both $\boldsymbol{\kappa}_{n}^{v}$ and $\boldsymbol{\kappa}_{n}^{a}$ are projected into joint embedding space via FC projection heads.
A cross-modal contrastive loss $\mathcal{L}_{cl}$ is then leveraged to align them with the instance discrimination supervision, so that they can learn cross-modal knowledge:
%Specifically, a cross-modal contrastive loss is leveraged to align $\boldsymbol{\kappa}_{n}^{v}$ and $\boldsymbol{\kappa}_{n}^{a}$ in a joint embedding space, so that they can learn cross-modal knowledge.
%First, both $\boldsymbol{\kappa}_{n}^{v}$ and $\boldsymbol{\kappa}_{n}^{a}$ are projected into a joint embedding space via FC block, as shown in Figure~\ref{fig:cmac}. 
%Then, the projected kernels $\boldsymbol{\kappa}_{n}^{v}$ and $\boldsymbol{\kappa}_{n}^{a}$ are constrained by:
\begin{equation}\label{eq:L_cl}\small
\mathcal{L}_{cl}(\boldsymbol{\kappa}_{n}^{v},\boldsymbol{\kappa}_{n}^{a})=-log \frac{sim(\boldsymbol{\kappa}_{n}^{v},\boldsymbol{\kappa}_{n}^{a})}{\sum_{m}sim(\boldsymbol{\kappa}_{n}^{v},\boldsymbol{\kappa}_{m}^{a})+\sum_{m\neq n}sim(\boldsymbol{\kappa}_{n}^{v},\boldsymbol{\kappa}_{m}^{v})}.
\end{equation}
Here we omit the projection heads for simplicity.
$\mathcal{L}_{cl}$ is a variant of contrastive loss, where the main difference between $\mathcal{L}_{cl}$ and $\mathcal{L}_{nce}$ is that we include within-modal negative pairs $(\boldsymbol{\kappa}_{n}^{v}, \boldsymbol{\kappa}_{m}^{v})$ in Eq.~\eqref{eq:L_cl} to introduce within-modal discrimination knowledge, which is ignored by most of the existing methods \cite{korbar2018cooperative,xiao2020audiovisual,patrick2020multi} as in Eq.~\eqref{eq:L_nce}. 
%which enables $f_v(\cdot)$ and $f_a(\cdot)$ to distinguish $\boldsymbol{\kappa}^{v}$ from $\boldsymbol{\kappa}^{a}$ in different videos clips.
%The main difference between $\mathcal{L}_{cl}$ and $\mathcal{L}_{nce}$ is that we include within-modal negative pairs $(\boldsymbol{\kappa}_{n}^{v}, \boldsymbol{\kappa}_{m}^{v})$ in Eq.~\eqref{eq:L_cl} to introduce within-modal knowledge, which is ignored by most of the existing methods \cite{korbar2018cooperative,xiao2020audiovisual,patrick2020multi} as in Eq.~\eqref{eq:L_nce}. 
AVID \cite{morgado2020audio} also considers the within-modal knowledge by adding $\mathcal{L}_{nce}(\boldsymbol{v}_{n1},\boldsymbol{v}_{n2})$ and $\mathcal{L}_{nce}(\boldsymbol{a}_{n1},\boldsymbol{a}_{n2})$ to Eq.~\eqref{eq:L_nce}, which, however, requires saving two clips for $n$-th video in a mini-batch as positive pairs.
In contrast, our simple yet effective modification requires no extra positive samples and no additional memory overhead.
%$\mathcal{L}_{cl}$ requires no extra positive samples and no additional memory overhead.

Notably, the contrastive learning of $\mathcal{L}_{cl}$ has two main effects on CMAC: a) introducing within- and cross-modal instance discrimination knowledge which has been proved crucial and fundamental for $f_v(\cdot)$ and $f_a(\cdot)$ \cite{korbar2018cooperative,patrick2020multi}; and b) bridging the modality gap between $\boldsymbol{\kappa}_{n}^{a}$, $\boldsymbol{\kappa}_{n}^{v}$ and $\boldsymbol{v}_{n}$, $\boldsymbol{a}_{n}$.

\noindent\textbf{Optimization.} Finally, the overall objective function of CMAC becomes:
\begin{eqnarray}\label{eq:L_all}
\begin{split}
\mathcal{L}_{all} =& \mathcal{L}_{cl}(\boldsymbol{\kappa}_{n}^{v},\boldsymbol{\kappa}_{n}^{a})+\mathcal{L}_{cl}(\boldsymbol{\kappa}_{n}^{a},\boldsymbol{\kappa}_{n}^{v})\\
&+\lambda[\mathcal{L}_{ac}(\boldsymbol{s}_{n}^{v},\hat{\boldsymbol{s}}_{n}^{v})+\mathcal{L}_{ac}(\boldsymbol{s}_{n}^{a},\hat{\boldsymbol{s}}_{n}^{a})],
\end{split}
\end{eqnarray}
where $\lambda$ is a hyper-parameter to balance contrastive loss and attention consistency loss.

\section{Experiments}
The experiments are conducted by transferring pretrained representations of Cross-Modal Attention Consistency (CMAC) to various downstream tasks, as well as several ablation studies.
For additional experimental results, please refer to the supplementary material.

\subsection{Pretraining Setting}
For pretraining, the standard visual-audio dataset of Kinetics-400 \cite{kay2017kinetics} is used, which contains $240$K videos of about 10 seconds.
After filtering out bad instances,~\eg no audio signals, about $220$K videos are used for pretraining.
The results on the large-scale AudioSet~\cite{gemmeke2017audio} are provided in supplementary material.

CMAC aims to learn a visual encoder $f_{v}(\cdot)$ and an audio encoder $f_{a}(\cdot)$ in unsupervised manner.
R(2+1)D-18 \cite{tran2018closer} is used as $f_{v}(\cdot)$, and ResNet-9 \cite{he2016deep} is used as $f_{a}(\cdot)$.
For each video instance in a batch, we randomly sample 1-second clip and produce visual modality data and audio modality data.
This makes our positive visual-audio pairs always synchronous from the same video and negative visual-audio pairs from different videos, which is different from \cite{korbar2018cooperative,patrick2020multi}.
The augmentation for the visual modality contains random cropping $128\times 128$ and random horizontal flipping.
The augmentation for audio modality employs log mel filtering with $257$ bank filters, time warping, random frequency masking, and random time masking in SpecAugment \cite{park2019specaugment}.
The contrastive learning in CMAC adopts MoCo \cite{he2020momentum} with default settings,~\emph{e.g.,} $\tau=0.07$, memory bank size as 15000, and momentum update parameter as 0.999.
The channels of cross-modal filters and contrastive head are $512$ and $256$, respectively.
For optimization, we use SGD optimizer with $lr=0.01$, weight decay $1e-5$, momentum $0.9$, and mini-batch size $128$ on 8 V100 GPUs.
A gradual warp-up schedule is utilized for the first 10 epochs.

\subsection{Downstream Setting}
\noindent\textbf{Video Action Recognition.} We evaluate the visual representations from $f_{v}(\cdot)$ on two widely-used benchmarks, UCF101 \cite{soomro2012ucf101} and HMDB51 \cite{kuehne2011hmdb}.
UCF101 contains \textasciitilde$13$K videos of 101 action classes, and HMDB51 contains \textasciitilde$7$K videos of 51 activity classes.
During training, we randomly sample $10$ clips of 32 frames for each video.
For visual augmentation, we follow \cite{patrick2020multi} and use random cropping, color jitter, and random horizontal flipping.
For optimization, we use SGD with $lr=0.005$, weight decay as $5e-3$, and momentum as 0.9.
We use a mini-batch of 16 and train for 12 epochs.
During testing, we uniformly sample 10 clips for each video, average their softmax scores, and report the class with the highest score.
The averaged top-1 accuracies across three split folds are reported for both UCF101 and HMDB51.

\noindent\textbf{Video Action Retrieval.} 
For the video retrieval task, we report the recall metric of 1, 5, 10 samples for split-1 of UCF101 and HMDB51 datasets, following the protocol in \cite{xu2019self}.
The features are directly extracted from the pretrained models without finetuning.
For each video, we uniformly sample 10 clips and average their pooled features after the last residual block.
We use samples from the validation set to query samples in the training set according to K-NN strategy and cosine distance.

\noindent\textbf{Audio Classification.}
We evaluate the audio representation from $f_{a}(\cdot)$ on ESC50 \cite{piczak2015esc}, DCASE2013 \cite{stowell2015detection} and DCASE2014 \cite{stowell2015detection} benchmarks by quickly training a linear classifier.
ESC50 contains $2$K audio clips from 50 different classes.
DCASE2013 and DCASE2014 contain 100 training clips from 10 different classes.
For each audio in ESC50, we sample 10 clips of 2-second length for training, and average the predicted scores of sampled 10 clips for testing.
For DCASE2013 and DCASE2014, we extract 60 clips of 1-second length for training and testing.

\subsection{Understanding Cross-Modal Attention Consistency}
We present four ablation studies to evaluate each component of CMAC.
For efficient validation, these experiments are conducted by pretraining on Kinetics with $50$ epochs, batch size of $64$, and finetuning on split-1 of UCF101 and HMDB51 without color jitter augmentation.

\subsubsection{Effect of each Component in CMAC}

CMAC essentially learns two kinds of knowledge, \ie, cross-modal attention consistency and instance discrimination, with $\mathcal{L}_{ac}$ and $\mathcal{L}_{cl}$, respectively.
Here, we evaluate their effects in Table \ref{tab:compoenent}.
We begin with the baseline of CMAC w/o $\mathcal{L}_{cl}$, since $\mathcal{L}_{cl}$ is also crucial to the attention consistency knowledge by bridging the modality gap between filters.
Note that removing both $\mathcal{L}_{cl}$ and $\mathcal{L}_{ac}$ will invalidate the pretraining (\ie from Scratch).
%Notably, the contrastive loss $\mathcal{L}_{cl}$ plays an important role in learning attention consistency knowledge by bridging the cross-modal filter gap.
%We begin with the baseline as CMAC w/o $\mathcal{L}_{cl}$. 
%Thus, we begin with the baseline as CMAC w/o FMC. Note that removing both FMC and PCF will invalidate the pretraining (\ie from scratch).
The results show that, without bridging the modality gap between filters, CMAC can hardly produce reasonable guided attentions for consistency preserving, thus leading to similar results to training from scratch.
Then, we add $\mathcal{L}_{cl}$ but remove $\mathcal{L}_{ac}$, which indicates that only instance discrimination knowledge is introduced.
This improves the performance from $76.4\%$ to $85.5\%$ on UCF101, demonstrating the great importance of cross-modal contrastive learning.
%cross-modal instance discrimination in this task.
We also list the inferior results of single modality contrastive learning for comparison.
%Besides, compared to instance discrimination within modality, it also proves that the cross-modal knowledge can boost the representation learning. 
Finally, with both $\mathcal{L}_{cl}$ and $\mathcal{L}_{ac}$, the result is further increased from $85.5\%$ to $87.2\%$, which proves the effectiveness and complementarity of attention consistency knowledge.
In summary, the bidirectional local correspondence between visual and audio signals can obviously benefit the fine-grained video content understanding. 

\begin{table}[t]
	\footnotesize
    \centering
    \caption{Evaluation of Attention Consistency (Att. Cons.) and Contrastive Loss (Contr. Loss) in CMAC.}
    \label{tab:compoenent}
    \vspace{0.2cm}
    \resizebox{1\columnwidth}{!}{
    \begin{tabular}{l|c|c|c|c}
	\shline
	Method&Att. Cons.&Contr. Loss&UCF101&HMDB51 \\
	\hline
	\hline
    Scratch &\xmark&\xmark&$73.2$&$23.7$ \\
    Single Modality&\xmark&\cmark&$82.5$&$47.3$ \\
	\shline
	CMAC w/o $\mathcal{L}_{cl}$ &\cmark&\xmark&$76.4$&$25.1$ \\
	CMAC w/o $\mathcal{L}_{ac}$ &\xmark&\cmark&$85.5$&$56.4$ \\
	CMAC &\cmark&\cmark& \textbf{$87.2$}&\textbf{$57.8$}\\
	\shline
    \end{tabular}}
\end{table}
\begin{figure}[t]
    \centering
    \includegraphics[width=1\columnwidth]{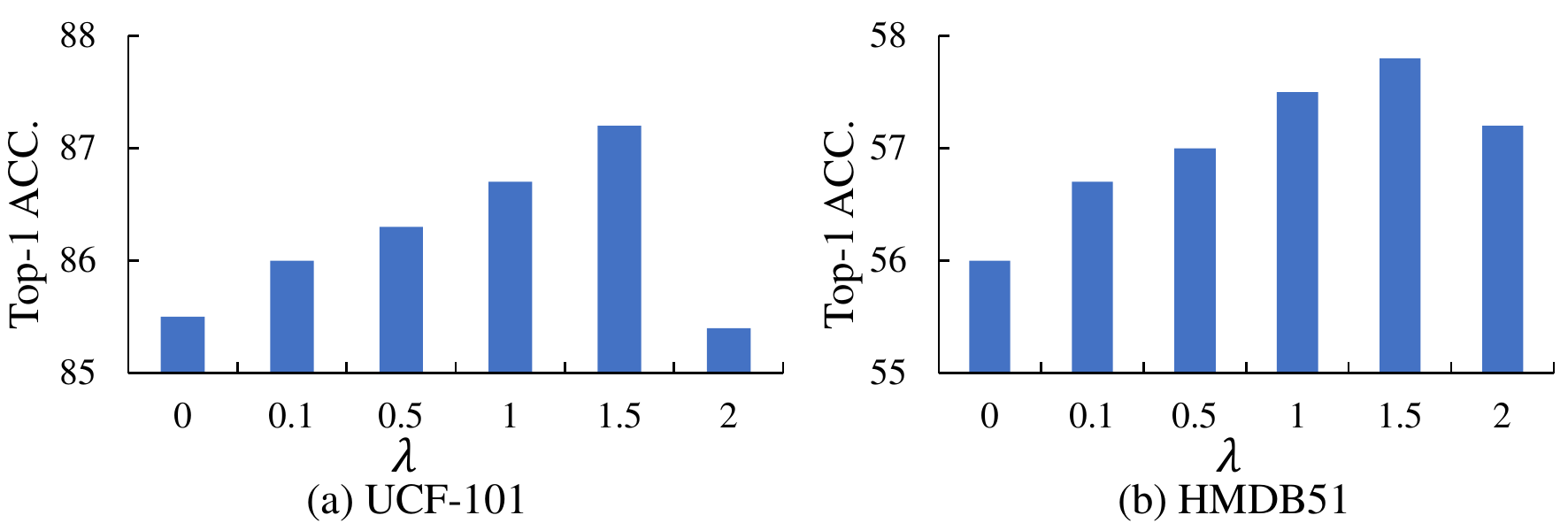}
    %\vspace{0.0cm}
    \caption{Evaluation for different $\lambda$ on UCF-101 and HMDB-51.}
    \label{fig:lambda}
    \vspace{-0.3cm}
\end{figure}

\subsubsection{Analysis of $\lambda$}
We evaluate the effects of cross-modal attention consistency $\mathcal{L}_{ac}$ by varying $\lambda$ in Eq.~\eqref{eq:L_all}.
From Fig.~\ref{fig:lambda}, it can be observed that, when $\lambda$ increases from $0$ to $1.5$, the performance of CMAC is stably improved from $85.5\%$ to $87.2\%$ on UCF101.
This proves that adding attention consistency for both visual and audio modalities can boost the encoders to understand visual content, thereby obtaining better downstream evaluation.
When $\lambda$ is larger than $1.5$, the performance drops.
The reason may be that the localized spatio-temporal regions from audio kernels are not accurate enough. Thus imposing strong cross-modal attention consistency may suffer from noisy attention maps.
Finally, we set $\lambda=1.5$ in the following experiments.

\subsubsection{Analysis of Pyramid Correlation Filtering}
In terms of pyramid correlation filtering (PCF), we conduct several experiments to search for a suitable architecture for generating the guided attention maps, as demonstrated in Table \ref{tab:simclr_cmsc}.
%to localize spatio-temporal regions and audio-spectrum.
The baseline is that PCF uses no normalization for convolution operation $(*)$ in Eq.~\eqref{eq:pcf} and no multi-scale fusion. 
Thus, the guided attention maps produced by filters and representations may be out of range $[0,1]$, which makes the attention gradient unstable.
Then, we add Softmax operation to normalize the response map, which brings gains on both datasets,~\eg $0.8\%$ improvement on UCF101.
However, the Softmax operation tends to suppress most of the regions, which slows the training at the beginning stage.
To this end, we further replace the Softmax with a cosine normalization for $*$, which brings obvious improvements.
Next, with the cosine normalization, we evaluate the effects of fusing multi-scale attention maps.
We observe that fusing two-scale attention maps obtains the best attention guidance for both datasets.
The reason may be that, due to limited memory resources, the spatial resolution of the final visual representation of $f_v(\cdot)$ has been reduced from $128\times 128$ to $8\times 8$, indicating that a two-scale fusion is enough. 

\begin{table}[t]
	\footnotesize
    \centering
    \caption{Evaluation of choices in the pyramid correlation filtering.}
    \vspace{0.2cm}
    %\resizebox{1\columnwidth}{!}{ 
    \begin{tabular}{l|c|c|c|c}
	\shline
	&$norm.$&Scale&UCF101&HMDB51 \\
	\hline
	\hline
	\multirow{5}{*}{\rotatebox{90}{PCF}}&None&1&$85.8$&$57.0$\\
	\cline{2-5}
	&softmax&1&$86.6$&$57.4$\\
	\cline{2-5}
	&\multirow{3}{*}{cosine}&1&$87.0$&$57.6$\\
	&&2&$87.2$&$57.9$\\
	&&3&$86.9$&$57.8$\\
	\shline
    \end{tabular}%}
    \label{tab:simclr_cmsc}
    \vspace{-0.3cm}
\end{table}

\begin{figure*}[h]
    \centering
    \includegraphics[width=2\columnwidth]{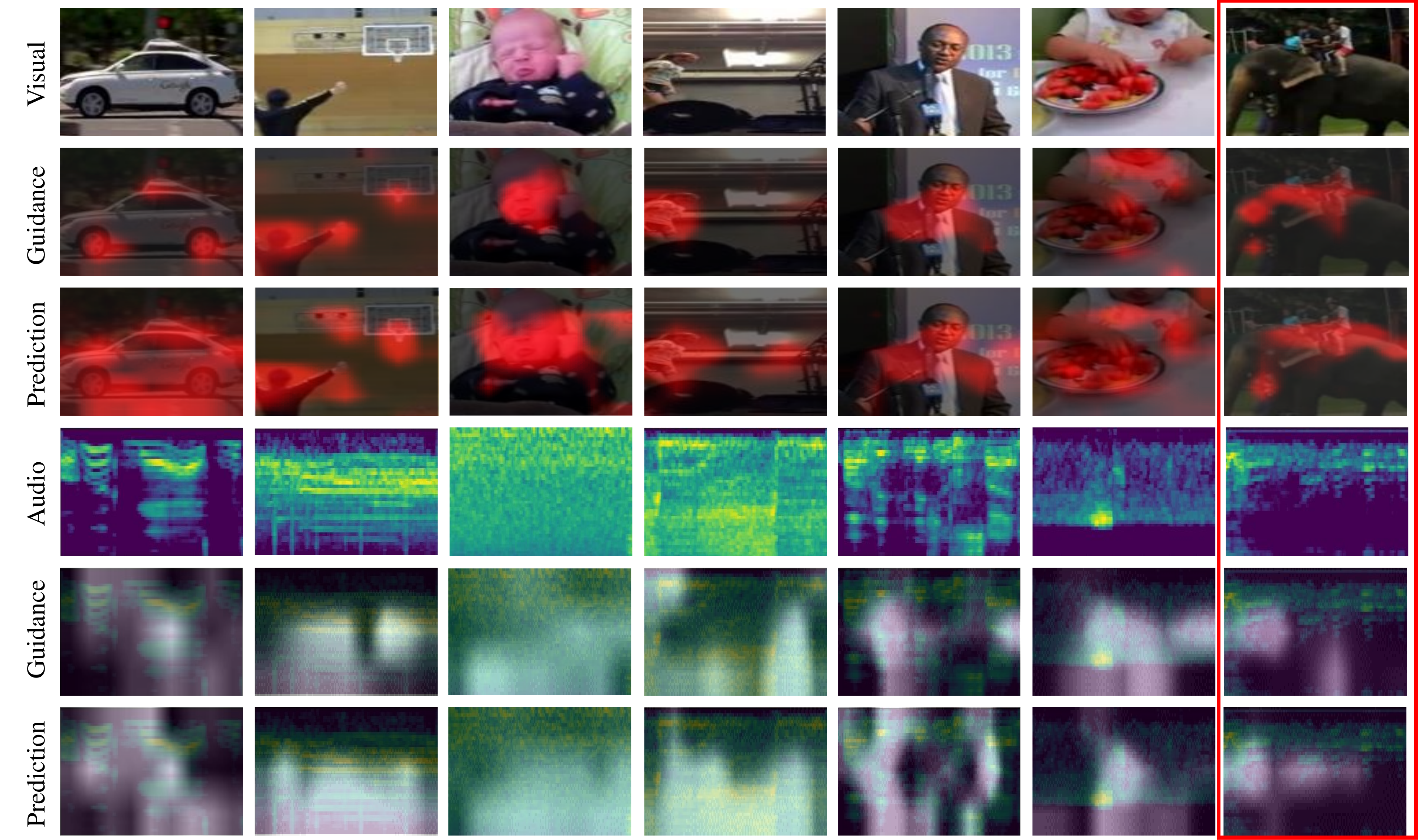}
    \vspace{0.2cm}
    \caption{Visualization of attention maps generated from cross-modal filters (Guidance) and encoders (Prediction) for both visual and audio modalities. The last column is a bad case with messy background and interlaced objects, where the localized regions are not precise.}
    \label{fig:vis}
\end{figure*}

\subsubsection{Analysis of Within-modal Information in Contrastive Loss}
Here we explore which kind of within-modal information really works in cross-modal contrastive learning.
Two types of sampling policies are considered: a) Within-modal negative sampling that takes two samples (with same modality) from different videos to form a negative pair and distinguishes them as in Eq. (\ref{eq:L_cl}); b) Within-modal positive sampling that takes two samples (with same modality) from the same video to form a positive pair and aligns them.
Note that using both policies actually forms the strategy in AVID \cite{morgado2020audio}.
%We split the within-modal information into two components: a) within-modal discrimination knowledge that repels the visual (audio) representations from different videos by introducing within-modal negative samples, as formulated in $\mathcal{L}_{cl}$; and b) within-modal consistency knowledge that attracts the visual (audio) representations from the same video by introducing within-modal positive samples, formulated in the supplementary material.
%Adding these two components actually makes up AVID \cite{morgado2020audio}.
The results are shown in Table~\ref{tab:cl_in_fmt}.
We observe that the negative sampling could benefit CMAC a lot on both datasets without extra memory expense, since we can simply utilize the samples from previous batch (stored in memory bank) to form negative pairs.
In contrast, the positive sampling harms the performance and even costs twice as much memory as negative sampling, due to that it requires saving two clips for each video. 
The possible reason for the performance drop is that aligning within-modal positive pairs would discard the intrinsic temporal synchronization in each video.

\red{Both the within-modal information in contrastive loss and the proposed attention consistency mechanism make up the impressive results of CMAC.
According to our observation, both of them can bring considerable improvements, indicating that they are complementary.}
%Considering the calculation efficiency and performance, we adopt batch size of $128$ plus within-modal negative sampling strategy as our setting.
\begin{table}[t]
	\footnotesize
    \centering
    \caption{Effects of within-modal negative samples (w/i Neg) and within-modal positive samples (w/i Pos) in the contrastive loss of CMAC. Iter. indicates the iteration number required for an epoch.}
    \vspace{0.2cm}
    %\resizebox{1\columnwidth}{!}{ 
    \begin{tabular}{l|c|c|c|cc}
	\shline
      &  w/i Neg & w/i Pos & Iter. & UCF101& HMDB51\\ 
	\hline
	\hline
    \multirow{4}{*}{CMAC}&\xmark&\xmark&$3,509$&$86.5$&$57.0$\\
    &\cmark&\xmark&$3,509$&$87.2$&$57.8$\\
    &\xmark&\cmark&$7,018$&$86.2$&$56.4$\\
    &\cmark&\cmark&$7,018$&$86.7$&$57.5$\\
	\shline
    \end{tabular}%}
    \label{tab:cl_in_fmt}
    \vspace{-0.3cm}
\end{table}

\begin{table*}[h]
	\footnotesize
    \centering
    \caption{Comparison of video action recognition on UCF-101 and HMDB-51, which are pretrained on Kinetics-400. Averaged top-1 accuracy across official splits is reported. Method with * indicates that additional video texts (\eg title) are used as supervision.}
    \vspace{0.2cm}
    %\resizebox{1.5\columnwidth}{!}{ 
    \begin{tabular}{l|l|l|c|c|c|c|c}
	\shline
     Methods&Backbone&Pretrained Dataset&~~~~Input Size~~~~&Batch Size&~~~Epoch~~~&~~UCF101~~&~HMDB51~\\
     \hline
     \hline
     %CBT\cite{sun2019learning}&S3D&Kinetics-600&$16\times 112^2$&128&267&79.5&44.6\\
     Multisensory\cite{owens2018audio}&3D-ResNet18&Kinetics-400&$64\times 224^2$&64&400&82.1&-\\
     CPD*\cite{li2020learning}&3D-ResNet50&Kinetics-400&$8\times 224^2$&128&110&88.7&57.7\\
     AVTS\cite{korbar2018cooperative}&MC3-18&Kinetics-400&$25\times 224^2$&64&90&85.5&56.9\\
     AV Sync+RotNet\cite{xiao2020audiovisual}&AVSlowFast&Kinetics-400&$64\times 224^2$&1024&120&87.0&54.6\\
     SpeedNet\cite{benaim2020speednet}&S3D&Kinetics-400&$16\times 224^2$&-&
81.1&48.8 \\     XDC\cite{alwassel2019self}&R(2+1)D-18&Kinetics-400&$32\times 224^2$&32&120&86.8&52.6\\
     AVID\cite{morgado2020audio}&R(2+1)D-18&Kinetics-400&$32\times 224^2$&256&400&87.5&60.8\\
     GDT\cite{patrick2020multi}&R(2+1)D-18&Kinetics-400&$32\times 128^2$&768&200&89.3&60.0\\
     \hline
     %CMAC&R(2+1)D-18&$32\times 128^2$&64&50&90.2&58.5\\
     CMAC&R(2+1)D-18&Kinetics-400&$32\times 128^2$&128&200&\textbf{90.3}&\textbf{61.1}\\
    \shline
    \end{tabular}%}
    \label{tab:vid_rec}
    \vspace{-0.3cm}
\end{table*}

\begin{table}[h]
	\footnotesize
    \centering
    \caption{Compasrison of full retrieval on UCF101 and HMDB51 datasets, which are pretrained on Kinetics-400.}
    \vspace{0.2cm}
    \begin{tabular}{l|l|l|l|l|l|l}
	\shline
	&\multicolumn{3}{c|}{UCF101}&\multicolumn{3}{c}{HMDB51} \\
	\cline{2-7}
	Recall @&1&5&10&1&5&10 \\
	\hline
	\hline
	%ST-Puzzle\cite{kim2019self}&19.7&28.5&33.5&-&-&-\\
	%OPN\cite{lee2017unsupervised}&19.9&28.7&34.0&-&-&- \\
	%ST Order\cite{buchler2018improving}&25.7&36.2&42.2&-&-&-\\
	ClipOrder\cite{xu2019self}&14.1&30.3&40.4&7.6&22.9&34.4\\
	SpeedNet\cite{benaim2020speednet}&13.0&28.1&37.5&-&-&-\\
	VCP\cite{luo2020video}&18.6&33.6&42.5&7.6&24.4&36.3 \\
	VSP\cite{cho2020self}&24.6&41.9&51.3&10.3&26.6&38.8 \\
	GDT\cite{patrick2020multi}&57.4&73.4&80.8&25.4&51.4&63.9 \\
	\hline
	CMAC&\textbf{58.4}&\textbf{74.2}&\textbf{81.3}&\textbf{26.0}&\textbf{52.2}&\textbf{64.2}\\
	\shline
    \end{tabular}
    \label{tab:retrieval}
    \vspace{-0.3cm}
\end{table}

\subsection{Visualization for Cross-modal Attention}
The key insight of CMAC is that forcing the visual encoder $f_{v}(\cdot)$ to attend to regions where sounds are made, and the audio encoder $f_{a}(\cdot)$ to ground the frequencies belonging to the interested objects.
Thus, we first visualize the attention maps generated from cross-modal filters in Fig.~\ref{fig:vis} (Guidance row).
It can be observed that, taking the car image as an example, our audio-guided attention can successfully localize the car wheel regions where the wheel noises are made in the visual frames.
%\begin{figure}[h]
%    \centering
%    \includegraphics[width=1\columnwidth]{figs/fig_bs.pdf}
%    \vspace{-0.1cm}
%    \caption{Evaluation of different pre-training batch size %of CMAC.}
%    \label{fig:bs}
%    \vspace{-0.2cm}
%\end{figure}
As a contrast, the attended regions from $f_{v}(\cdot)$ itself (Prediction row) focus on the whole car body, because $f_{v}(\cdot)$ is more interested in global information without cross-modal local correspondence.
It reveals that the visual guidance from audio signals could be an effective supervision to constrain $f_v(\cdot)$ to focus on local regions with sounds, which brings significant improvements as shown in Table~\ref{tab:compoenent}.
In terms of audio encoder $f_a(\cdot)$, the localized frequencies with guidance from visual signals are more sparse than the focused frequencies from $f_a(\cdot)$ itself.

However, CMAC is not always effective in localizing correct visual and audio local regions, as shown in the last column of Fig.~\ref{fig:vis}.
When the video has the messy background and interlaced objects, it is hard to precisely match the corresponding local patterns without human annotations.
Nevertheless, these observations prove that the cross-modal attention consistency is intuitive and reasonable in visual-audio unsupervised representation learning.

\subsection{Comparison with State-of-the-arts}
Given our best setting, we train CMAC with longer epochs of 200 and strong color jitter augmentation in \cite{patrick2020multi}, and transfer the pretrained representations to various downstream tasks.

\noindent\textbf{Video Recognition.} We evaluate CMAC on two video action recognition benchmarks of UCF101 and HMDB51.
The results are reported in Table~\ref{tab:vid_rec}, which shows that CMAC obtains impressive results on both UCF101 and HMDB51.
It should be noted that CMAC is a computation-friendly method.
Particularly, GDT \cite{patrick2020multi} utilizes $6$ times of GPU cost and batch size than CMAC, while CMAC surpasses GDT about $1.0\%$ on UCF101 and $1.1\%$ on HMDB51.
Compared to AVID \cite{morgado2020audio}, CMAC also adopts much smaller batch size and fewer epochs to obtain $2.8\%$ improvement on UCF101 and $0.3\%$ on HMDB51.
The results reveal that the bidirectional local correspondence is effective in boosting the visual representation learning.
After teaching the visual encoder where to focus on, it becomes easier to understand the visual content and distinguish different video instances.
Thus, CMAC requires fewer batch size and epoch to surpass most of the existing methods.

\noindent\textbf{Video Retrieval.} 
For video retrieval, we evaluate the pretrained representations from CMAC on UCF101 and HMDB51 without finetuning the backbone.
The results are given in Table~\ref{tab:retrieval}, from which we observe that CMAC achieves new state-of-the-art performance in all settings.
This shows that video representations of similar object content from CMAC become closer than those from previous methods, because the attention consistency makes the visual encoder focus on the correct object regions. 

\begin{table}[t]
\centering
% \vspace{-1em}
\captionsetup[subfloat]{captionskip=2pt}
\captionsetup[subffloat]{justification=centering}
\scriptsize
\caption{Comparison of audio classification by quickly training a linear classifier. The pretrained  dataset is Kinetics-400.}
\label{tab:audio}
\vspace{-1em}
\subfloat[$*$ indicates that the backbone encoder $f_{a}(\cdot)$ is also finetuned on ESC50.
\label{tab:audio:ESC50}]{
%\tablestyle{1.2pt}{1}
\resizebox{0.45\columnwidth}{!}{
\begin{tabular}{l|c}
	\shline
	Methods&~~ESC50~~ \\
	\hline
	\hline
	%Piczak ConvNet\cite{piczak2015environmental}&-&$64.5$&$-$\\
	%Ensemble\cite{stowell2015detection}&-&$-$&$77$\\
    SoundNet\cite{aytar2016soundnet}~~~~&$74.2$\\
    AVTS\cite{korbar2018cooperative}&$76.7$\\
    AVID\cite{morgado2020audio}&$79.1$\\
	XDC*\cite{alwassel2019self} &$78.0$ \\
	GDT\cite{patrick2020multi} & $78.6$  \\
	\hline
	CMAC &\textbf{81.4}\\
	\shline
\end{tabular}}
    }\hspace{2mm}
\subfloat[The baseline is a general cross-modal contrastive loss as defined in Eq.~\eqref{eq:L_cmcl}.]{
\tablestyle{1.2pt}{1}
\resizebox{0.4\columnwidth}{!}{
\begin{tabular}{l|c|c}
	\shline
	Methods&D2013&D2014 \\
	\hline
	\hline
	%Piczak ConvNet\cite{piczak2015environmental}&-&$64.5$&$-$\\
	%Ensemble\cite{stowell2015detection}&-&$-$&$77$\\
    Scratch&48&-\\
    Single-Modal~~~&61&-\\
    \hline 
    Baseline& 68&89\\
    AVTS\cite{morgado2020audio}&-&91 \\
	GDT\cite{patrick2020multi}& $73$&94  \\
	\hline
	CMAC &\textbf{76}&\textbf{96}\\
	\shline
\end{tabular}}
}
\vspace{-0.3cm}
\end{table}

\noindent\textbf{Audio Classification.} 
For the audio classification task, we evaluate the pretriained model from CMAC on ESC50, DCASE2013, and DCASE2014 benchmarks.
Notably, the reported results of compared methods are pretrained on Kinetic-400.
Table~\ref{tab:audio} demonstrates that the pretrained model from CMAC outperforms existing methods by a large margin,~\eg $2.8\%$ improvement on ESC50.
Note that XDC further finetunes the backbone $f_a(\cdot)$ on the downstream task, while CMAC only learns the classifier and obtains better performance.
For DCASE2013 and DCASE2014, the result of GDT is obtained by using their official model pretrained on Kinetics-400.
The superiority attributes to the localized voice frequencies according to target objects, which also proves the effectiveness of the proposed CMAC.

\section{Conclusion}
In this paper, we propose a novel pretext task for visual-audio unsupervised representation learning, namely Cross-Modal Attention Consistency (CMAC).
The core insight of CMAC is that the visual perception encoder should attention to regions where sounds are made, and the auditory perception should ground acoustic-frequencies of sounding objects, which we call bidirectional local correspondence.
To model such a bidirectional local correspondence supervision, CMAC aims to align the regional attention maps purely from visual signals with the target attention guidance from audio signals, and vice versa.
%devises a pyramid correlation filtering to align the regional salience maps purely from visual signals with the target attention guidance from audio signals, and vice versa.
Accompanied by a remoulded cross-modal contrastive loss with additional within-modal interactions, CMAC obtains impressive results on various downstream tasks.
%In further, we will further explore the temporally local correspondence between adjacent visual frames and adjacent acoustic-frequencies to capture fine-grained video supervision.

{\small
\bibliographystyle{ieee_fullname}
\bibliography{egbib}
}

\end{document}